\documentclass[letterpaper, 10 pt, conference]{ieeeconf}  
\usepackage[T1]{fontenc}
\usepackage{xcolor}
\usepackage{graphicx}
\usepackage{amsmath}
\usepackage{glossaries}
\usepackage{dblfloatfix}

\usepackage{fancyhdr}

\fancypagestyle{withfooter}{
    \fancyhead[C]{\textcolor{gray}{This paper has been accepted for publication in the Proceedings of the 2024\\IEEE/RSJ International Conference on Intelligent Robots and Systems (IROS 2024)}}
    \setlength{\headheight}{15pt}  

}

\DeclareMathOperator*{\ODESolve}{ODESolve}
\DeclareMathOperator*{\sgn}{sgn}

\newcommand{\pare}[1]{\left({#1}\right)}

\newcommand{\state}{x}
\newcommand{\dstate}{\dot{\state}}
\newcommand{\vel}{v}
\newcommand{\acc}{a}
\newcommand{\vx}{\vel_x}
\newcommand{\vy}{\vel_y}
\newcommand{\yawRate}{r}
\newcommand{\engineSpeed}{\omega_s}

\newcommand{\yawRateEst}{\hat{\yawRate}}
\newcommand{\vxEst}{\hat{\vel}_x}
\newcommand{\vyEst}{\hat{\vel}_y}

\newcommand{\imuVel}{\omega^{IMU}}
\newcommand{\imuAcc}{\acc^{IMU}}
\newcommand{\imuAccX}{\imuAcc_x}
\newcommand{\imuAccY}{\imuAcc_y}
\newcommand{\steeringAngle}{\delta}
\newcommand{\motorCurrent}{I_q}

\newcommand{\measurements}{y}
\newcommand{\measurementFunction}{h}
\newcommand{\stateEst}{\hat{\state}}
\newcommand{\control}{u}
\newcommand{\dvx}{\dot{\vel}_x}
\newcommand{\dvy}{\dot{\vel}_y}
\newcommand{\dvxEst}{\hat{\dot{\vel}}_x}
\newcommand{\dvyEst}{\hat{\dot{\vel}}_y}
\newcommand{\model}{f}
\newcommand{\dstateEst}{\hat{\dstate}}

\newcommand{\frictionCoeff}{\mu}
\newcommand{\frictionCoeffA}{\frictionCoeff_A}
\newcommand{\frictionCoeffB}{\frictionCoeff_B}
\newcommand{\frictionCoeffC}{\frictionCoeff_C}
\newcommand{\frictionCoeffD}{\frictionCoeff_D}
\newcommand{\frictionCoeffEst}{\hat{\frictionCoeff}}

\newcommand{\accEst}{\hat{\acc}}
\newcommand{\accEstX}{\accEst_x}
\newcommand{\accEstY}{\accEst_y}

\newcommand{\transsmisionForceLoss}{\tau_t}
\newcommand{\transsmisionKC}{k_{tC}}
\newcommand{\transsmisionKV}{k_{tv}}

\newcommand{\mass}{m}
\newcommand{\Fdrag}{F_{drag}}

\newcommand{\slipRatio}{\kappa}
\newcommand{\slipAngle}{\alpha}

\newcommand{\Bx}{B_x}
\newcommand{\Cx}{C_x}
\newcommand{\Dx}{D_x}
\newcommand{\Ex}{E_x}
\newcommand{\By}{B_y}
\newcommand{\Cy}{C_y}
\newcommand{\Dy}{D_y}
\newcommand{\Ey}{E_y}

\newcommand{\tireModel}{F}
\newcommand{\Fxr}{F_{x_r}}
\newcommand{\Fxf}{F_{x_f}}
\newcommand{\Fyr}{F_{y_r}}
\newcommand{\Fyf}{F_{y_f}}

\newcommand{\Loss}{\mathcal{L}}

\newcommand{\processCov}{R}
\newcommand{\measurementCov}{Q}
\newcommand{\processCovL}{L_\processCov}
\newcommand{\measurementCovL}{L_\measurementCov}

\newcommand{\PC}{\text{PC}}
\newcommand{\PCR}{\text{PCR}}
\newcommand{\NN}{\text{NN}}
\newcommand{\NNTire}{\text{NNT}}
\newcommand{\NNTireMu}{\text{NNTF}}
\newcommand{\UKF}{\text{U}}
\newcommand{\Hetero}{\text{H}}

\newacronym[plural=MDPs, firstplural=Markov Decision Processes (MDPs)]{mdp}{MDP}{Markov Decision Process}
\newacronym{ekf}{EKF}{Extended Kalman Filter}
\newacronym[plural=UKFs, firstplural=Unscented Kalman Filters (UKFs)]{ukf}{UKF}{Unscented Kalman Filter}
\newacronym[plural=IMUs, firstplural=Inertial Measurement Units (IMUs)]{imu}{IMU}{Inertial Measurement Unit}
\newacronym[plural=RNNs, firstplural=Recursive Neural Networks (RNNs)]{rnn}{RNN}{Recursive Neural Network}
\newacronym[plural=GRUs, firstplural=Gated Recurrent Units (GRUs)]{gru}{GRU}{Gated Recurrent Unit}
\newacronym{lstm}{LSTM}{Long Short-Term Memory}
\newacronym{esc}{ESC}{Electronic Speed Controllers}
\newacronym[plural=GNSSs, firstplural=Global Navigation Satellite Systems (GNSSs)]{gnss}{GNSS}{Global Navigation Satellite System}
\newacronym[plural=OVSs, firstplural=Optical Velocity Sensors (OVSs)]{ovs}{OVS}{Optical Velocity Sensor}
\newacronym{pemm}{PEMM}{Prediction Error Minimisation Method}

\newacronym{mse}{MSE}{Mean Squared Error}
\newacronym{mae}{MAE}{Mean Absolute Error}
\newacronym{99ae}{99\%-AE}{99th percentile of Absolute Error}

\newacronym{pem}{PEM}{Prediction Error Method}

\newacronym{mlp}{MLP}{Multilayer Perceptrons}

\IEEEoverridecommandlockouts                              

\title{\LARGE \bf
Learning dynamics models for velocity estimation in autonomous racing 
}

\author{Jan W\k{e}grzynowski, Grzegorz Czechmanowski, Piotr Kicki and Krzysztof Walas
\thanks{All authors are with IDEAS NCBR, Warsaw, Poland and with Institute of Robotics and Machine Intelligence, Poznan University of Technology, Poznan, Poland {\tt\small name.surname@ideas-ncbr.pl}}%
}

\begin{document}

\maketitle
\thispagestyle{withfooter}

\begin{abstract}
Velocity estimation is of great importance in autonomous racing. Still, existing solutions are characterized by limited accuracy, especially in the case of aggressive driving or poor generalization to unseen road conditions. To address these issues, we propose to utilize \gls{ukf} with a learned dynamics model that is optimized directly for the state estimation task. Moreover, we propose to aid this model with the online-estimated friction coefficient, which increases the estimation accuracy and enables zero-shot adaptation to the new road conditions. To evaluate the \gls{ukf}-based velocity estimator with the proposed dynamics model, we introduced a publicly available dataset of aggressive manoeuvres performed by an F1TENTH car, with sideslip angles reaching~40\textdegree. Using this dataset, we show that learning the dynamics model through \gls{ukf} leads to improved estimation performance and that the proposed solution outperforms state-of-the-art learning-based state estimators by 17\% in the nominal scenario. Moreover, we present unseen zero-shot adaptation abilities of the proposed method to the new road surface thanks to the use of the proposed learning-based tire dynamics model with online friction estimation.
\end{abstract}

\section{INTRODUCTION}
In recent years, we have observed a huge interest in self-driving technologies, starting from driver assistance systems and ending with fully autonomous vehicles. Implementing these technologies reduces transportation costs and air pollution, and in the long run, gives a promise of reduced traffic and much safer and more accessible transportation~\cite{autonomouscarsreport}. Most of the research in this area focuses on calm maneuvers that cover most of the driving time. However, from the safety perspective, it is especially important to develop systems prepared for everyday scenarios and emergencies that may require aggressive maneuvers with high sideslips to avoid a collision. 

\begin{figure}[t]
\centering
\includegraphics[width=\linewidth]{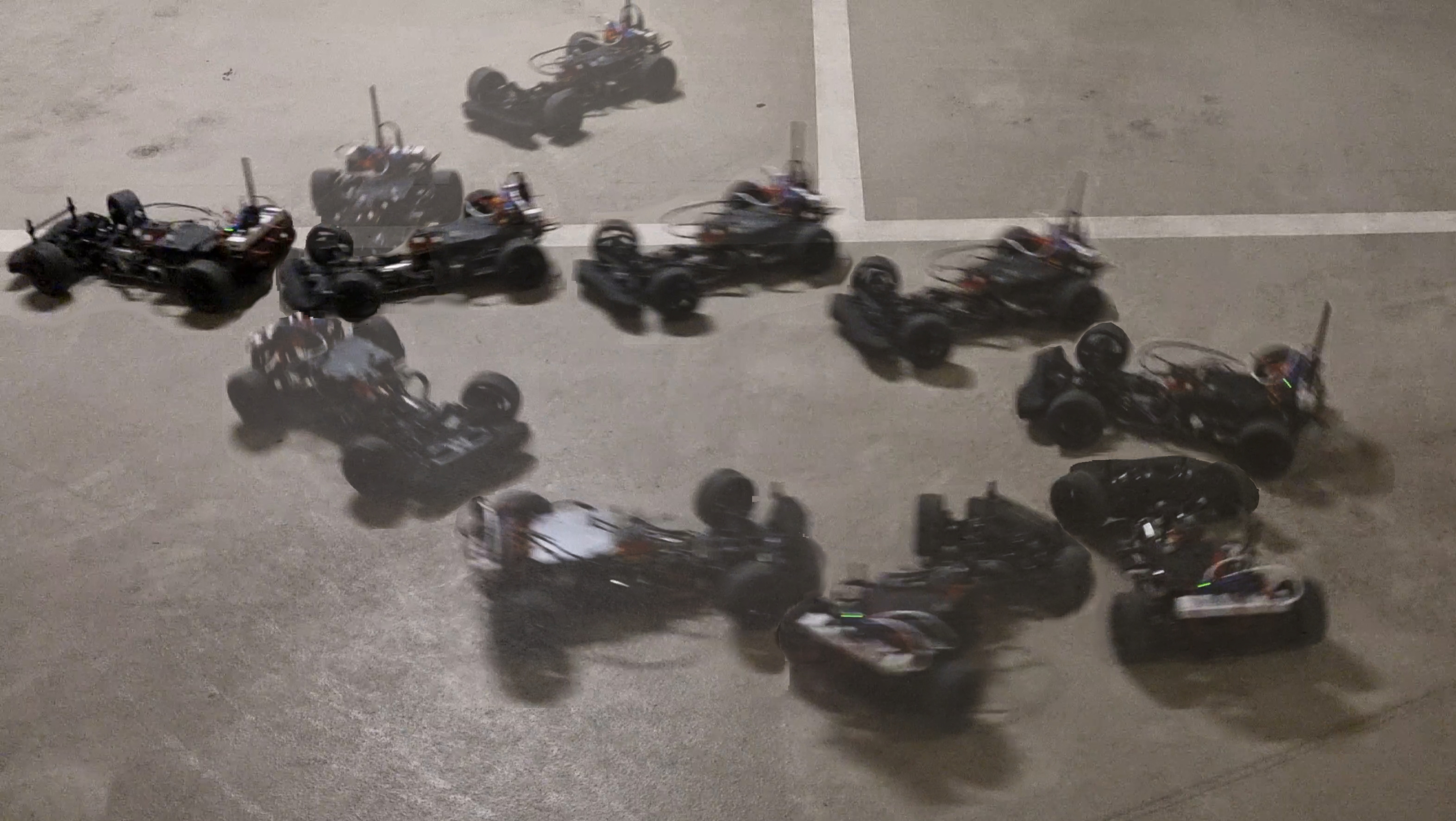}
\vspace{-0.6cm}
\caption{The main goal of this paper is to learn a dynamics model of an F1/10 car for accurate velocity estimation during aggressive maneuvers.}
\vspace{-0.5cm}
\label{fig:frontpage}
\end{figure}

An application that exploits aggressive maneuvers on a daily basis is autonomous racing (see Fig.~\ref{fig:frontpage}). Thus, we want to focus on it in this paper. In particular, we tackle the problem of velocity estimation, which is an important part of both perception and control pipelines, as it allows us to undistort the LiDAR measurements and provide state feedback. While it is possible to use exteroceptive sensors, like GNSS or ground speed sensors, to measure the velocity, they increase the cost of the platform and are prone to failure or bias in challenging environments. Therefore, we focus on the problem of velocity estimation using only proprioception, namely \gls{imu} and motor speed.

One of the most popular approaches to this problem relies on using Kalman Filtering, especially \gls{ekf} ~\cite{wischnewski2019ekf, gosala2019ekfracing} and \gls{ukf}~\cite{kim2020sideslip}~\cite{bertipaglia2023ukfsideslip}. These methods strongly rely on the accuracy of the car dynamics model, which is crucial in the parts of the state space in which the considered system is hard to observe. This can be addressed by using the hybrid car dynamics models, which consist of the physical car model aided with a neural network to enable accurate dynamics modelling~\cite{chee2023nodeukf, kim2022physicsad}. However, these models are trained to estimate the state derivative accurately among the whole state space, which may not be optimal from the state estimation point of view.
It is also possible to utilize fully data-driven state estimators like recursive neural networks (RNNs)~\cite{srinivasan2020endtoend, ghosn2023lstm} as they allow for highly accurate and flexible approximation of the car and measurement dynamics. However, pure machine learning-based methods are typically limited in terms of generalization abilities, like adjusting to changing tire-road friction~\cite{lampe2023neuralfriction}. 

To address the aforementioned issues, we propose to exploit the concept of differentiable Kalman filters~\cite{haarnoja2016backpropkf, kloss2021howtotrain} in the process of learning the dynamics model of a racing car. Although this approach is not new, it is mostly used to train measurement models~\cite{kloss2021howtotrain, liu2023denkf, buchnik2023latentkf} and its impact on learning data-driven dynamics models of the complex robotics systems for state estimation is not well studied~\cite{bai2023survey}. Surprisingly, learning a dynamics model through a differentiable Kalman filter is said to have no impact on the state estimation accuracy~\cite{kloss2021howtotrain}. Still, this paper argues the opposite based on the thorough experimental analysis. In fact, we claim that learning the dynamics model in this way follows a better optimization criterion, as opposed to training the model to be an accurate one-step predictor, and allows the neural network-based model to focus on the accurate prediction along the more important state space dimensions from the estimation point of view.
Moreover, we show that in the considered case learning an accurate car model is more important for the prediction step than for the observation.

Moreover, we exploit the structure of the \gls{ukf} to estimate not only the car velocities but also the tire-road friction, which allows the learned state estimator to adapt to new road conditions without the need to collect new data and retrain. This zero-shot test time adaptability is crucial in deploying reliable state estimators in the real world, which is typically the main drawback of machine learning-based solutions.
Finally, we analyze the sources of the improved velocity estimation performance while learning the dynamics model through \gls{ukf}.

To compare the proposed approach with state-of-the-art state estimation approaches and validate the aforementioned claims, we introduce a dataset of a very dynamic driving of a 1:10 scale F1 car model~\cite{f1tenth}. This allows us to safely examine the state estimation in severe conditions, including huge side-slip, slip ratios, lateral accelerations, and decreased tire-road friction.

Our contributions can be summarized as follows:
\begin{itemize}
    \item We show that learning the car dynamics model through \gls{ukf} increases the state estimation accuracy and allows for outperforming state-of-the-art estimators based on \glspl{rnn}, while maintaining an interpretability of \gls{ukf} estimates.
    \item We propose to aid a hybrid of learning-based and analytical car dynamics models with an online estimated tire-road friction, which increases the estimation accuracy and enables adaptation to the new conditions during test time in a zero-shot manner.
    \item We introduce a publicly available 52 minutes long dataset of the aggressive maneuvers performed by a F1TENTH car~\cite{f1tenth} (see~Fig.~\ref{fig:frontpage}) with side-slip angles reaching 40\textdegree, which is far beyond what one can find in the publicly available datasets.
\end{itemize}

\section{RELATED WORK}
In this section, we will discuss related papers both from the perspective of the considered application, i.e., velocity estimation of a car during aggressive maneuvers, and the approach we propose, i.e., machine learning-based car dynamics modeling and online tire-road friction estimation.

One of the most classic yet effective approaches to car velocity estimation utilizes the \gls{ekf}~\cite{gosala2019ekfracing, wischnewski2019ekf} that fuses the data from several sensors and uses the car dynamics model to estimate the lateral velocity. In turn, the advancements in the field of machine learning inspired many researchers to incorporate neural networks into the Kalman filtering pipeline. The authors of~\cite{kim2020sideslip}~and~\cite{bertipaglia2023ukfsideslip} used a neural network to perform as a virtual sideslip sensor, which measurements are then filtered out using \gls{ukf}. In~\cite{kim2020sideslip} the neural network is trained to predict the sideslip in a supervised manner based on the data from the simulation, while in~\cite{bertipaglia2023ukfsideslip} the real data is used and the loss is computed based on the output of the \gls{ukf}, similarly like in this paper. A different approach was introduced in~\cite{escoriza2021kalmannetvelest}, where the \gls{rnn} is utilized to predict the Kalman gain, while the rest of the pipeline is not taking advantage of the representational power of neural networks.

The racing car velocity estimation task can also be solved without using Kalman filtering but by exploiting the \glspl{rnn} trained in an end-to-end manner to maintain some latent representation of the robot state and estimate the velocities out of it~\cite{srinivasan2020endtoend, ghosn2023lstm}.
A \glspl{gru} based \gls{rnn} has been utilized in~\cite{srinivasan2020endtoend} and allowed for an accurate estimation of the vehicle velocity during maneuvers with high sideslip (10$^{\circ}$  at the rear axle) and slip ratio ($ \approx 20\%$), without using external velocity sensors. 
Authors of~\cite{ghosn2023lstm} furthered this research by using \gls{lstm} units to process the sensory information and fully connected layers to estimate the velocity based on the previous estimates and the hidden state of the \gls{lstm}. This approach led to superior accuracy on a real-world, full-scale car dataset compared to traditional model-based and other learning-based methodologies. Nonetheless, a significant limitation of RNN-based systems is their lack of built-in mechanisms for estimating uncertainty, a critical component for ensuring safety during aggressive driving maneuvers. Moreover, due to the end-to-end machine learning-based solutions, these models are potentially more susceptible to the \textit{out-of-distribution} data, which is essential for safety across diverse road conditions.

To make the machine learning-based state estimators more robust, some recent works focused on enhancing the learning-based models with some physical structure~\cite{chee2023nodeukf, lio2023kinematic}. 
In~\cite{lio2023kinematic}, the neural network used to predict the state estimate is fitted into a kinematic structure of the vehicle to satisfy kinematic laws. However, it lacks insights into the vehicle's dynamics. 
In turn, in~\cite{chee2023nodeukf}, like in this paper, an \gls{ukf} with a hybrid vehicle dynamics model is considered. They enhance the classical model with a residual neural ordinary differential equation and train them to predict the state derivative that best matches the next observation. 
Our paper shows that learning residual dynamics model along the classical one lead to similar results as for fully-learned dynamics, both having problems with generalization.
Instead, we propose to aid the classical dynamics model with a neural network modeling the tire-forces, like in~\cite{kim2022physicsad} or~\cite{djeumou2023autonomous}. Authors of~\cite{kim2022physicsad} follow the structure of the Pacejka tire model and use a neural network to predict its parameters, while authors of~\cite{djeumou2023autonomous} introduce an \textit{ExpTanh} tire model. 
In turn, we do not impose this kind of structure on the tire model but use the neural network to model the tire forces almost directly. 

One of the main determinants of the forces generated by the tires is the tire-road friction coefficient~\cite{matusko2008friction}. There were several attempts to learn how to predict it in real-world scenarios using neural networks~\cite{lampe2023neuralfriction} trained \textit{end-to-end} or hybrid models~\cite{matusko2008friction, lowenstein2023pimhe} adapted online based on the prediction errors. We rather follow the second approach, but instead of adapting some friction predictor, we estimate the friction coefficient online using \gls{ukf}, and we use it to scale the neural network tire model output.
Moreover, we show that training the dynamics model through the \gls{ukf} leads to notably better results in terms of the state estimation performance than simply training the dynamics model to be an accurate one-step predictor as it is typically proposed in literature.

\section{MATERIALS AND METHODS}

\subsection{Problem definition}
In this paper, we consider a problem of the estimation of the vehicle state $\state$ defined by
\begin{equation}
   \state = [\vx,\vy, \yawRate, \engineSpeed], 
\end{equation}
where $\vx$ and $\vy$ are longitudinal and lateral velocities in the local vehicle frame, $\yawRate$ is a yaw rate, and $\engineSpeed$ is the engine speed scaled by wheel radius and gear ratio.
We assume that the estimation should be performed using (i) the proprioceptive sensors that almost every robot is equipped with, i.e. \gls{imu}, which gives us linear accelerations $\imuAcc$ and angular velocities $\imuVel$, and velocity measurements $\engineSpeed$ from the electric motor driver, as well as (ii) control signals consisting of steering angle $\steeringAngle$ and motor current $\motorCurrent$.

The considered task is not trivial due to the aggressive nature of the racing driving style that includes both longitudinal and lateral slips and, therefore, limited observability of the system state. Because the system state is not measurable directly, a reliable and precise system model is necessary to obtain the state estimates from time series of measurements and control inputs. Moreover, we would like to consider the case in which the parameters of the environment are non-stationary and non-uniform, so we are not making popular assumptions about the uniform tire-road friction coefficient over the whole driving area and driving time. In fact, we want to challenge the state estimators to generalize to different friction conditions, preferably without tedious data collection and retraining.

Note that a similar problem to the one presented above is typically solved by incorporating the \gls{imu} measurements directly into the simplified dynamics model~\cite{wischnewski2019ekf, gosala2019ekfracing}. However, this approach is not feasible in the problem considered in this paper due to the lack of an external velocity sensor (line GNSS or optical velocity sensor) that can correct the drift caused by the integration of measured acceleration.

\subsection{Proposed solution}
\begin{figure}[t]
    \centering
    \includegraphics[width=\linewidth]{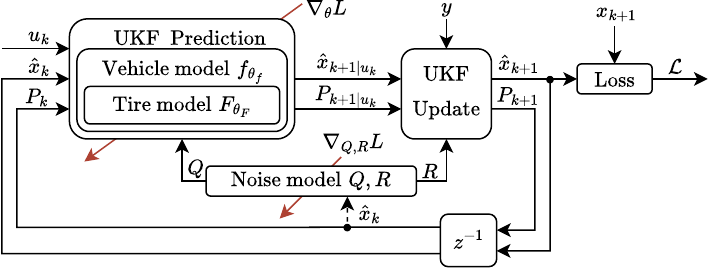}
    \vspace{-0.5cm}
    \caption{The general scheme of the differentiable \gls{ukf}-based state estimation, with learnable vehicle and noise models. We propose to train the car dynamics and noise models through \gls{ukf} to minimize the state estimation error.}
    \label{fig:diagram}
    \vspace{-0.3cm}
\end{figure}

To address the problem introduced in the previous section, we propose to use an \gls{ukf} framework~\cite{wan2000ukf}, as it is a standard approach to state estimation of nonlinear systems. In particular, we want to exploit the differentiable implementation of \gls{ukf}~\cite{kloss2021howtotrain} to learn an accurate hybrid (understood as a combination of classical and learning-based) model of the system dynamics that will allow us to estimate the robot state  
during aggressive driving precisely. The general scheme of the proposed approach is presented in Fig.~\ref{fig:diagram}. The presented scheme consists of the classical UKF prediction and update steps, with tunable vehicle and noise models based on the gradient of the estimation loss function. In the next subsections, we describe these elements in detail.

\subsubsection{UKF update}
In the update step, following the standard \gls{ukf} algorithm~\cite{wan2000ukf}, we compute the difference between the sensor measurements $\measurements$ and the output of the measurement function $\measurementFunction(\stateEst, \control)$, and scale it by the Kalman gain to obtain a state estimate update. 
The yaw rate $\yawRate$ and wheels speed $\engineSpeed$ are directly measured and comparable with their estimates. However, to compare with longitudinal $\imuAccX$ and lateral $\imuAccY$ accelerations from \gls{imu} we need to compute the estimated accelerations in the inertial frame by $\accEstX = \dvxEst - \yawRateEst \vyEst$ and $\accEstY = \dvyEst + \yawRateEst \vxEst$, where $\dvxEst, \dvyEst$ are the local acceleration estimates based on the dynamics model $\dstateEst = \model(\stateEst, \control)$ and $\yawRateEst, \vxEst, \vyEst$ are the estimates of the $\yawRate, \vx$ and $\vy$.

\subsubsection{UKF prediction}
\label{sec:ukf_prediction}

In the prediction step, the system dynamics model estimates the future state based on the current state and control inputs. This smooths out the information obtained from update steps and adjusts it to comply with the dynamics of the observed system. 
The accuracy of the dynamics model is particularly important in those parts of the state space where access to reliable measurements is limited. Still, in our specific case, its importance is even greater because it is also a key part of the measurement model.

In this paper, we analyze the whole family of different vehicle dynamics models, ranging from single track car with Pacejka tire model to the fully-learned vehicle model, which are inspired by the approaches from literature~\cite{wischnewski2019ekf, nagabandi2018nnmodel, xu2022tireforcesnn}, and extend them in a more structured way of learning the vehicle dynamics.

\textbf{Single track model} We generally consider the vehicle dynamics as a continuous function $\dstate = \model(\state, \control)$, which can be integrated to obtain the next discrete state.
In our considerations, we omit simplistic dynamics models, like a point mass or kinematic bicycle, as they are not feasible in the case of dynamic driving. Instead, we employ a single-track model for its accurate representation of the basic vehicle dynamics, which can be defined by
\begin{equation}
\resizebox{.91\hsize}{!}{$
\begin{split}
\dvx &= \frac{1}{m} \pare{\Fxr + \Fxf\cos(\delta) - \Fyf \sin(\delta) - \Fdrag + \mass \vy \yawRate},\\
\dvy &= \frac{1}{m} \pare{\Fxf \sin(\delta) + \Fyr +\Fyf \cos(\delta) - \mass \vx \yawRate},\\
\dot{r} &= \frac{1}{I_z} \pare{\pare{\Fxf \sin(\delta) + \Fyf \cos(\delta)} l_f - \Fyr l_r}, \\
\dot{\omega}_{s} &= \frac{1}{I_e} \pare{k_{\phi} \motorCurrent - R \cdot \Fxf - R \cdot \Fxr - \transsmisionForceLoss},
\label{eq:single_track}
\end{split}$}
\end{equation}
where $\mass$ is the vehicle mass, $\Fdrag$ is the aerodynamic drag force acting of the vehicle, assumed to scale quadratically with $\vx$, $\delta$ is the steering angle, $I_z$ is the vehicle moment of inertia around the $z$-axis, $l_r$ and $l_r$ represents distances from center of mass to rear and front axis respectively, $R$ is the wheel radius, $I_e$ is powertrain inertia, $k_{\phi}$ is the motor's torque constant, while $\Fxr, \Fxf, \Fyr, \Fyf$ represents the $x$ (longitudinal) and $y$ (lateral) components of the lumped tire forces generated by the rear and front virtual wheels. 
We assume a DC motor model with torque proportional to motor current $\motorCurrent$. 
This torque is reduced by the transmission system losses, which are modeled as $\transsmisionForceLoss = \transsmisionKC \sgn(\engineSpeed) + \transsmisionKV \engineSpeed$ acting on the motor shaft, where $\transsmisionKC$, $\transsmisionKV$ represents Coulomb and viscous friction coefficient respectively.

\textbf{Pacejka tire model} The most challenging part of \eqref{eq:single_track} is how to model the lumped forces that model the complex phenomenon of the tire-road interaction. One of the most popular approaches to this matter is to express these forces using Pacejka Magic Formula~\cite{pacejka}
\begin{equation}
   \begin{split}
    F_x &= \frictionCoeff \Dx \sin \left(\Cx \arctan \left(\Bx \kappa \right. \right. \\
    &\quad \left. \left. - \Ex \left(\Bx \kappa - \arctan(\Bx \kappa) \right) \right) \right),\\
    F_y &= \frictionCoeff \Dy \sin \left(\Cy \arctan \left(\By \slipAngle \right. \right. \\
    &\quad \left. \left. - \Ey \left( \By \slipAngle - \arctan(\By \slipAngle) \right) \right) \right).
\end{split}
\label{eq:pacejka_tire_model}
\end{equation}
where $\Bx, \Cx, \Dx, \Ex, \By, \Cy, \Dy, \Ey$ represents the parameters of the model for longitudinal and lateral lumped tire forces, $\frictionCoeff$ is the friction coefficient between the tire and driven surface, $\slipRatio$ is the slip ratio and $\slipAngle$ is the slip angle. Note that we use the same parameters to obtain the longitudinal forces acting on the front or rear tire, but in the case of lateral force, the $\By, \Cy, \Dy, \Ey$ parameters can be different for the front and rear virtual wheels. Similarly, the slip angle $\slipAngle$ is computed separately for specific wheels.

\textbf{Neural car model} Tire forces, commonly represented with \eqref{eq:pacejka_tire_model}, offer good generalization but may not adequately capture complex tire behaviors due to their limited flexibility. 
This issue led to the development of data-driven car models that typically utilize neural networks to model vehicle dynamics. One of the most straightforward approaches in this spirit~\cite{nagabandi2018nnmodel} is to let the neural network approximate the whole dynamics 
\begin{equation}
    \dstate = \model_{NN}(\state, \control).
    \label{eq:full_nn_model}
\end{equation}
However, despite neural networks' great capacity to surpass classical models in making predictions on data similar to their training environment, their applicability to new scenarios is limited, as they offer limited interpretability and generalization.

\textbf{Neural tire model} To address these issues, we propose a hybrid approach that integrates analytically derived models with neural networks, aiming to achieve both generalizability and accuracy. Analytical models address simpler behaviors, while neural networks tackle the more complex aspects, such as tire forces~\cite{xu2022tireforcesnn, djeumou2023autonomous}. 
In our solution, similarly to~\cite{xu2022tireforcesnn}, a neural network is applied to represent all tire forces 
\begin{equation}
    [\Fxr, \Fxf, \Fyr, \Fyf] = \tireModel_{NN}(\state, \control),
\label{eq:nn_tire_model}
\end{equation}
while the vehicle dynamics is expressed by \eqref{eq:single_track}. 

\textbf{Friction-scaled neural tire model} To enhance the adaptability and interpretability of the neural tire model \eqref{eq:nn_tire_model}, we employ the friction coefficient $\frictionCoeff$ to scale tire forces generated by the neural network such that
\begin{equation}
    [\Fxr, \Fxf, \Fyr, \Fyf] =  \frictionCoeff \cdot \tireModel_{NN}(\state, \control).
\label{eq:nn_friction_tire_model}
\end{equation}
This allows us to adjust the predictions of the neural tire model to the driving conditions that may not be present during training just by knowing the estimate of the friction coefficient $\frictionCoeffEst$ during the test. In particular, we exploit the fact that we are using the \gls{ukf} framework and we use it to estimate the friction coefficient online, which is described in more detail in Section~\ref{sec:firction_coeff_est}.

\subsubsection{Noise model}
\label{sec:noise_models}
A crucial component of Bayesian estimation methods is the noise model~\cite{greenberg2023okf}. It quantifies the uncertainty of the process and measurement models using covariance matrices $\processCov$ and $\measurementCov$. 
In this paper, we omit the diagonal noise model due to its oversimplification.
Instead, we incorporate the off-diagonal elements into the covariance matrices and define the noise model by
\begin{equation}
    \processCov = \processCovL  \processCovL^T + I \cdot \epsilon , \quad \measurementCov = \measurementCovL  \measurementCovL^T + I \cdot \epsilon,
    \label{eq:nois_model_formulation}
\end{equation}
where $L_R$ and $L_Q$ are lower triangular matrices with trainable elements, $I$ is the identity matrix, and $\epsilon$ is small number (e.g. 1e-7) which enhances the numerical stability.
This formulation guarantees that both $\processCov$ and $\measurementCov$ matrices remain positive definite throughout the training process.

    
The noise model introduced above assumes that the covariance matrices are constant, which may not be true, e.g., the noise on the \gls{imu} may increase when the car is moving very fast on uneven surfaces. To account for these kinds of effects, we utilize a heteroscedastic noise model, which has been shown to significantly impact \gls{ukf}-based state estimation~\cite{kloss2021howtotrain}. In this noise model, we predict the elements of the $\processCovL$ and $\measurementCovL$ matrices based on the current state estimate $\stateEst$ using linear regression and compute $\processCov$ and $\measurementCov$ using \eqref{eq:nois_model_formulation}.

\vspace{0.2cm}
\subsubsection{Model training procedure}
\label{sec:training_procedure}

All the dynamics models described in Section~\ref{sec:ukf_prediction}, whether classically derived from first principles or exploiting neural networks, are parameterized with $\theta$. The values of $\theta$ have a tremendous impact on the dynamics model accuracy, so they need to be identified or learned prior to the use of the model in the state estimator. Thanks to the differentiability of the considered models, one can optimize their parameters $\theta$ using gradients $\nabla_\theta \Loss$ of some loss function $\Loss$ w.r.t $\theta$.

To learn an accurate dynamics model $\model(\state, \control)$, one must decide on the loss function $\Loss$ formulation. The most popular way is to train the model to minimize the square of the one-step prediction error $e_k = (\state_{k+1} - \ODESolve(\model(\state_k, \control_k), T_s))^2$, where $T_s$ is the step duration and $\ODESolve$ is a numerical integrator, which in our case is the 4-th order Runge-Kutta method.
This way, the model is optimized to accurately predict the local state derivative $\dstate$ such that its integral fits the measured evolution of the system state $\state_{k+1} - \state_k$.

Although training dynamics models to be accurate one-step predictors is a well-established approach, it does not consider the context of their application.
This would make no difference for perfect models, but in case of their limited capacity, it may be suboptimal from the specific task point of view. 
Therefore, we propose to utilize the differentiable implementation of \gls{ukf}~\cite{kloss2021howtotrain} and optimize the parameters of the dynamics model, such that the resultant \gls{ukf} achieves the smallest possible state estimation errors instead of the minimal one-step prediction errors.
This way, we optimize a better criterion from the final task point of view.

Besides learning the dynamics model through \gls{ukf}, one can also use the same procedure to update the parameters of the noise models, i.e., $\processCovL$ and $\measurementCovL$ for the homoscedastic process and observation noise models or parameters of the heteroscedastic noise model (see Section~\ref{sec:noise_models}). The importance of this step
is thoroughly described in~\cite{greenberg2023okf}. 

Moreover, we observe that simultaneous learning of the dynamics model and noise parameters is also an important factor in the final estimation performance, as it allows for the deliberate adaptation of the noise models to the changing dynamics model.
In particular, in the case of the state estimation, we may not need a dynamics model that makes small prediction errors on average for the whole state space. 
Instead, it may prefer to focus on some key components of the state space that are hard to be measured, and compensate for the resulting prediction errors of the remaining components by reducing their corresponding measurement covariances.


Following the training methods described above, the dynamics model training can be separated into two parts:
    \begin{itemize}
        \item pretraining based on the one-step prediction loss minimized on the batch of samples,
        \item estimation fine-tuning, where noise model and prediction model parameters are tuned through the differentiable implementation of \gls{ukf} on sequences of samples by minimizing the \gls{mse} between ground-true states and state estimates (see Fig.~\ref{fig:diagram}).
    \end{itemize}
This way, we can quickly pretrain an accurate dynamics model and then adapt it to the desired task of velocity estimation. While it is possible to train the model through \gls{ukf} from scratch, we found that way inefficient and prone to being stuck in local minima.

\subsubsection{Friction coefficient estimation}
\label{sec:firction_coeff_est}

In Section~\ref{sec:ukf_prediction}, in particular in \eqref{eq:nn_friction_tire_model}, we proposed to enhance the tire dynamics model by scaling the output of the neural network-based model by the friction coefficient $\frictionCoeff$. In many papers regarding the racing car velocity estimation, the tire-road friction coefficient is assumed to be constant, e.g.~\cite{bertipaglia2023ukfsideslip, djeumou2023autonomous}, which seems quite impractical taking into account possible changes of the weather conditions or wear out of the tires. Alternatively, it may be treated as a disturbance~\cite{srinivasan2020endtoend, ghosn2023lstm}, to which the prediction model needs to become robust, which may lead to suboptimal estimation performance and require collecting the data with a wide spectrum of $\frictionCoeff$. In turn, in~\cite{lampe2023neuralfriction}, only several specific friction coefficients are considered and assumed to be constant and homogeneous on large track areas.

To overcome these simplifying assumptions we propose to estimate $\frictionCoeff$ online exploiting the \gls{ukf} framework, by adding $\frictionCoeff$ to the state vector
\begin{equation}
        \state = [\vx, \vy, \yawRate, \engineSpeed, \frictionCoeff].
\label{eq:state_def_with_friction}
\end{equation}
As we cannot model the dynamics of the friction coefficient, we assume it to be constant $\dot{\frictionCoeff} = 0$, but at the same time, we allow the adjustments of $\frictionCoeff$ during the update step of the \gls{ukf}, based on the observed discrepancy between the observed and predicted measurements. 

\begin{table*}[!b]
\centering
\vspace{-0.2cm}
\caption{Comparison of estimation performance on data from the tire used during training}
\vspace{-0.2cm}
\begin{tabular}{l|ccccccccccc}
{Method} &  $\PC$ & $\PCR$ \cite{chee2023nodeukf} & $\NN$ \cite{nagabandi2018nnmodel} &  $\NN_{\UKF}$ & $\NNTire$ & $\NNTire_{\UKF}$ & $\NNTireMu$ &  $\NNTireMu_{\UKF}$ & $\NNTireMu_{\UKF\Hetero}$ & GRU~\cite{srinivasan2020endtoend}  & LSTM~\cite{ghosn2023lstm} \\ \hline
{MSE} & 0.0654 & 0.0089 & 0.0078 & \textbf{0.0058} & 0.0088 & 0.0087 & 0.0085 & \textbf{0.0072} & \textbf{0.0063} & 0.0076 & 0.0338 \\
{MAE} & 0.1459 & 0.0543 & 0.0558 & \textbf{0.0464} & 0.0578 &  0.0564 & 0.0574 &  0.0539 & \textbf{0.0493 }& 0.0524 & 0.1277 \\ 
{99\%-AE} & 0.5893 & \textbf{0.2512} & \textbf{0.2385} & \textbf{0.1681} & \textbf{0.24} & \textbf{0.2444} & \textbf{0.2184} & \textbf{0.1777} & \textbf{0.1743} & 0.2608 & 0.3825 \\   
\end{tabular}
\label{tab:test_same_surface}
\end{table*}

\subsection{Dataset}
\label{sec:dataset}
The development of state estimation algorithms requires the availability of high-quality, diverse data. Unfortunately, previous studies \cite{srinivasan2020endtoend, ghosn2023lstm, escoriza2021kalmannetvelest} often did not publish datasets, limiting the ability to benchmark and improve estimation methods. The most relevant publicly available dataset for developing agile estimation algorithms is Targa Sixty-Six \cite{TargaSixtySix}, which contains data from a Ferrari 250 LM driven at Palm Beach International Raceway at a speed reaching 200 km/h. However, even though they feature vehicles traveling at speeds exceeding 220 km/h, they do not extensively cover high side-slip maneuvers that occur while operating vehicles at the limits of handling. This notable gap in the availability of challenging, real-world datasets has propelled us to develop and publicly release our dataset.

The dataset introduced in this paper is created from real-world data obtained using an F1TENTH vehicle~\cite{f1tenth} and an external motion capture system. The experimental vehicle, an Xray GTXE'22, is a four-wheel drive, 1/8th scale RC car. It weighs 4.5 kg and is powered by a single motor controlled by a motor driver. It captures raw sensor data at varying frequencies: the \gls{imu} at 400Hz and the motor RPM, motor q-axis current, and steering angle at 100Hz. High-precision positional and orientation data was captured using an OptiTrack motion capture system at 100 Hz. Data collection utilized two computers (one onboard for sensor data and another for motion capture) synchronized with Chrony to minimize time discrepancies. To facilitate accurate state estimation, these raw measurements were linearly interpolated and sampled at 100 Hz. Ground truth vehicle velocities were calculated by differentiating OptiTrack position data with a Savitzky-Golay filter \cite{savitzky-golay}, using a second-order polynomial and an 8 ms filter window. The velocities were then sampled at the same timestamps as the raw measurements.

\begin{figure}[t]
    \centering
    \includegraphics[width=\linewidth]{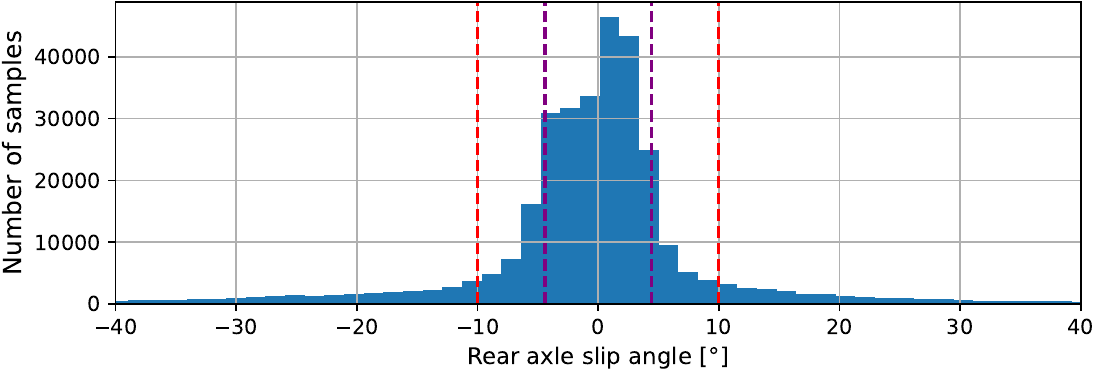}  
    \vspace{-0.6cm}
    \caption{Distribution of the rear axle slip angle in the introduced dataset. Maximum value reported in \cite{srinivasan2020endtoend} (red) and 99th Percentile from Targa Sixty-Six dataset \cite{TargaSixtySix} (purple) are added as reference.}
    \label{fig:dataset}
    \vspace{-0.2cm}
    \end{figure}

This way, we collected 52 minutes of driving time with a remotely controlled car. Thanks to this, we were able to collect a diverse set of high slip and high acceleration maneuvers, for which the 99th percentile (to ignore outliers) of the slip angle at the rear axle exceeds maximum values reported in previous works \cite{srinivasan2020endtoend, ghosn2023lstm, TargaSixtySix} by a substantial margin (see Fig.~\ref{fig:dataset}). 
In fact, our dataset features dynamic maneuvers with a slip angle at the rear axle surpassing 40\textdegree  (99th percentile), which presents a much more demanding estimation challenge than existing datasets.

Moreover, typically, the racing datasets contain data for specific tire-road conditions and neglect the complexity of real-life racing, e.g., tire wear during the race or wet asphalt because of the rain.
To take such circumstances into account and facilitate the research in the direction of robust data-driven state estimators we increased the diversity of the introduced dataset by using 4 different sets of tires, each offering different coefficients of friction - tire A (20.1\% of the dataset, $\frictionCoeffA \approx 0.65$), tire B (18.7\%, $\frictionCoeffB \approx 0.58$), tire C (50.2\%, $\frictionCoeffC \approx 0.43$) and tire D (11\%, $\frictionCoeffD \approx 0.45$). Tires C and D are meant to simulate the conditions of the decreased friction, e.g., due to the wet road surface, which we obtained by covering tires A and B with two different types of tape.

\section{EXPERIMENTS}

In this section, we examine the performance of the \gls{ukf} with dynamics models introduced in Section~\ref{sec:ukf_prediction} and relate it to the state-of-the-art state estimation methods based on \glspl{rnn}~\cite{srinivasan2020endtoend, ghosn2023lstm}.
To do so, we use the dataset introduced in Section~\ref{sec:dataset}.

\subsection{Experimental setup}
We consider the following dynamics models: 
\begin{enumerate}
    \item $\PC$ -- a classical single track vehicle model \eqref{eq:single_track} with Pacejka tire model~\eqref{eq:pacejka_tire_model},
    \item $\PCR$ -- $\PC$ model with learned residual dynamics \cite{chee2023nodeukf},
    \item $\NN$ -- a fully neural network-based vehicle model~\eqref{eq:full_nn_model} as proposed in~\cite{nagabandi2018nnmodel},
    \item $\NNTire$ -- a classical single track vehicle model \eqref{eq:single_track} with neural network-based tire model~\eqref{eq:nn_tire_model} inspired by~\cite{xu2022tireforcesnn},
    \item $\NNTireMu$  -- a classical single track vehicle model \eqref{eq:single_track} with the proposed neural network-based tire model scaled by online-estimated friction coefficient~\eqref{eq:nn_friction_tire_model}.
\end{enumerate}
All of these models are used in the \gls{ukf} framework and are trained to minimize the one-step prediction loss for 1000 epochs. To evaluate the impact of learning through the \gls{ukf}, as described in Section~\ref{sec:training_procedure}, we fine-tune some models for another 1000 epochs on the 500 steps-long state estimation problems (see Section~\ref{sec:seq_len}), and denote it with subscript~$_{\UKF}$.
Validation and testing were conducted using 10s sequences.
Moreover, in all cases, we optimize the parameters of the noise models through \gls{ukf}~\cite{greenberg2023okf}.
All trainings were performed with a learning rate of $5\cdot10^{-4}$.
For all models, the default noise model is homoscedastic with full crosscovariance matrix~\eqref{eq:nois_model_formulation}, while the models with heteroscedastic noise are denoted by the subscript $_{\Hetero}$ (see Section~\ref{sec:noise_models}).
Neural network tire models employ \gls{mlp} that consist of three layers, with each layer comprising 64 neurons, where $\PCR$ and $\NN$ are designed using also 3 layer \gls{mlp} with 256 neurons each.

To further extend the set of the baseline methods, we implemented \gls{rnn}-based state estimators presented in~\cite{srinivasan2020endtoend} -- GRU and \cite{ghosn2023lstm} -- LSTM.
We used the same neural networks and training procedure as detailed in these papers.
Since GRU and Robust LSTM models require time to stabilize the hidden states for accurate estimations, we adopted the recommendation from \cite{srinivasan2020endtoend}, and we neglect the first 200 samples (2s, marked with gray in Fig.~\ref{fig:ukf_rnn_comparison}) in both training and evaluation. 
This technique ensures the hidden states and estimates of the models are well-formed, allowing a fair comparison.
    
To obtain a single estimation loss value from all of the state variables, we weigh the states by the inverse of their variability, calculated as the mean absolute difference of subsequent samples, which, for our dataset, after normalizing to 1 are equal to 0.223, 0.506, 0.157, and 0.114 for $\vx, \vy, \yawRate,$ and $\engineSpeed$ respectively.
Note, that \gls{rnn}-based approaches estimate only the lateral, longitudinal, and rotational velocity, which ground true values are provided by a motion capture system.
On the other hand, the dynamics of wheel speed is an important aspect in the whole vehicle motion description and must be modeled in our \gls{ukf} estimator.
Thus for evaluation purposes, we set the weight of $\engineSpeed$ to 0.

\begin{table*}[t]
    \centering
    \caption{Comparison of estimation performance for the tire unseen during training tire}
    \vspace{-0.2cm}
    \begin{tabular}{l|cccccccccccc}
    {Method} & $\PC$ & \PCR \cite{chee2023nodeukf} & \NN \cite{nagabandi2018nnmodel} & $\NN_{\UKF}$ &  $\NNTire$ & $\NNTire_{\UKF}$ & $\NNTireMu$ &  $\NNTireMu_{\UKF}$ & $\NNTireMu_{\UKF\Hetero}$ & $\NNTireMu_{\UKF\Hetero}^{*}$ & GRU~\cite{srinivasan2020endtoend} &  LSTM~\cite{ghosn2023lstm} \\ \hline
    {MSE}    & 0.489 & 0.0769 & 0.0770 & 0.0781 & -& - & 0.0311 & 0.0259 & \textbf{0.0237} & \textbf{0.0188} & 0.0967 & 0.1328 \\
    {MAE}   &  0.255 & 0.159 & 0.154 & 0.162 & - & - & 0.105 & 0.0991 & \textbf{0.0939} & \textbf{0.0810} & 0.1811 & 0.2224 \\
    {99\%-AE} & 2.621 & 0.758 & 0.763 & 0.734 & - & - & 0.596 &  0.553 & \textbf{0.504} & \textbf{0.377} & 0.8735 & 1.0165 \\ 
    \end{tabular}
    \label{tab:test_new_surface}
    \vspace{-0.5cm}
\end{table*}

\subsection{Velocity estimation in known conditions}


To evaluate our method against existing approaches, we utilized data from a single tire type (A), dividing it into training, validation, and test sets with proportions of 70\%, 20\%, and 10\%, respectively. 
After training, we evaluated all considered approaches on the test set and reported \gls{mse}, \gls{mae}, and \gls{99ae} in Tab.~\ref{tab:test_same_surface}.

Our proposed method $\NNTireMu_{\UKF\Hetero}$ significantly outperformed both baseline models \cite{srinivasan2020endtoend, ghosn2023lstm} in terms of all considered metrics, improving results of \gls{gru} by 17\%, 6\% and 33\% for \gls{mse}, \gls{mae}, and \gls{99ae} respectively. The results show that each of the introduced components, i.e., online estimated friction coefficient, learning through \gls{ukf} and heteroscedastic noise model, play their role in achieving state-of-the-art performance.
Moreover, one can see that in terms of all considered learning-based dynamics models, learning through \gls{ukf} improves the estimation accuracy, which shows the importance of adapting the model to the target task. Interestingly, even when tested in the same conditions as in the training, \gls{rnn}-based models exhibit poor worst-case performance - higher \gls{99ae} than \gls{ukf}-based state estimators with learned dynamics models.

It is worth noting that, in general, the best performance was achieved by a fully neural network-based vehicle model trained through \gls{ukf}. 
However, as we will show next, this kind of end-to-end dynamics model generalizes poorly to the data outside of the training distribution.

\subsection{Zero-shot generalization to new road conditions}

Velocity estimation for an autonomous vehicle is a safety-critical task that must remain reliable under changing road conditions.
Thus, we evaluate the method's capability for zero-shot generalization in new tire/road friction scenarios.
To do so, we use models trained on tire A and evaluate them using the tire C dataset (33\% smaller $\frictionCoeff$).
The results of this experiment are presented in Tab.~\ref{tab:test_new_surface}.
Due to shifts in the data distribution, \gls{ukf} with $\NNTire$ model become unstable, which we denote with '$-$.' For the remaining methods, we observe a notable drop in the estimation performance. However, it is the smallest for the proposed $\NNTireMu$ models. These results prove its unparalleled zero-shot generalization abilities to the new road conditions, as $\NNTireMu_{\UKF\Hetero}$ achieves 3 and 4 times smaller \gls{mae} than $\NN$ and \gls{gru} respectively.
Notably, even better results may be achieved when a prior knowledge about $\frictionCoeff$ is available. If we assume that initial friction coefficient estimate $\frictionCoeffEst$ is equal to $0.4$, instead of $0.6$, the tracking performance of the proposed method further increases (see $\NNTireMu_{\UKF\Hetero}^{*}$ in Tab.~\ref{tab:test_new_surface}).

\begin{figure}[t]
    \centering
    \includegraphics[width=\linewidth]{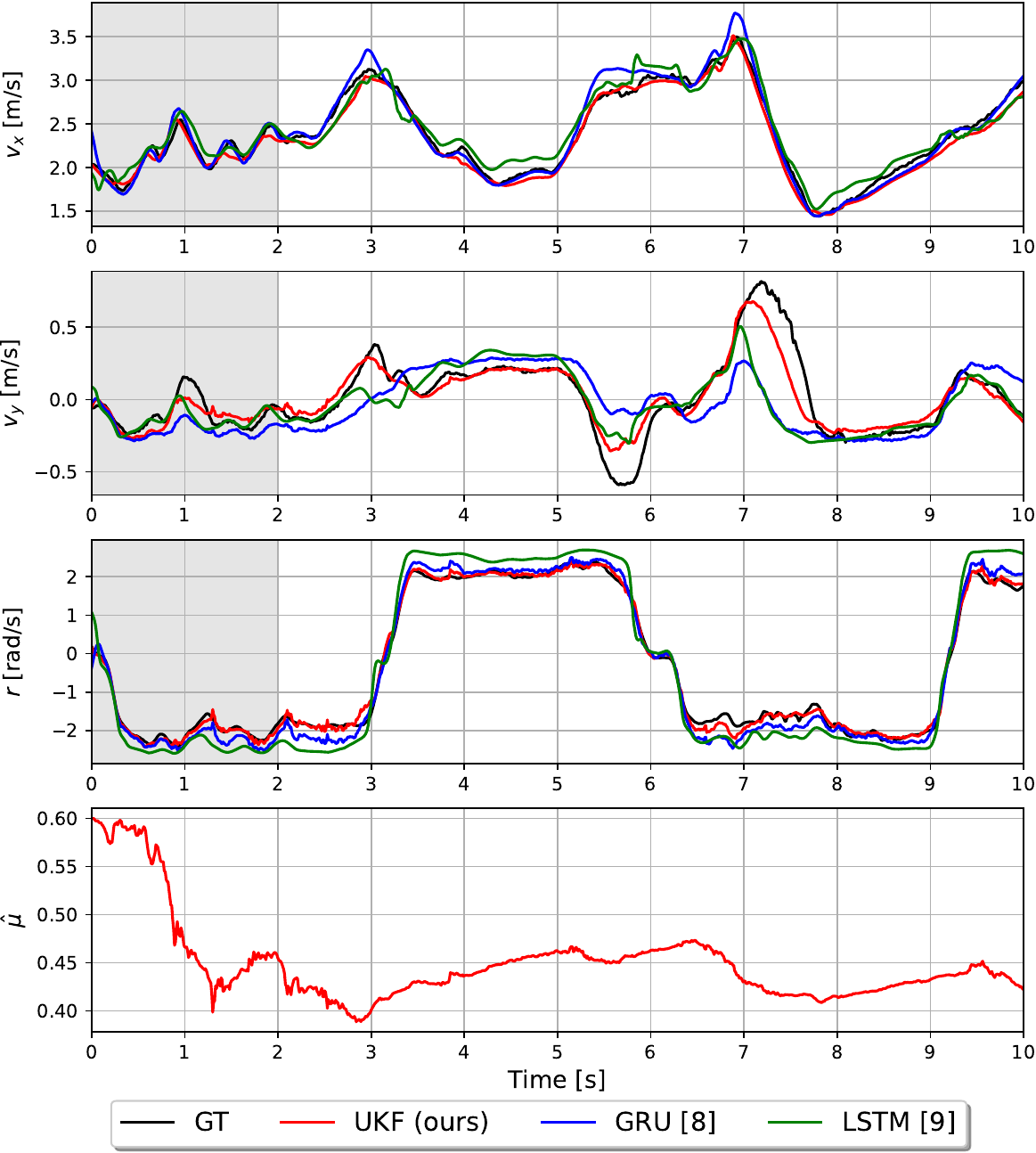}
    \vspace{-0.6cm}
    \caption{Estimation performance on tire C - unseen during training.}
    \label{fig:ukf_rnn_comparison}
\end{figure}

In Fig.~\ref{fig:ukf_rnn_comparison}, we present a sample challenging test sequence, which shows the differences in the velocity estimation between the considered solutions and the evolution of the friction coefficient estimate $\frictionCoeffEst$. One can see that the proposed solution tracks the car velocities very accurately and loses the track only in extremely hard conditions, i.e., high side-slip (5.7s) and simultaneous hard braking and turning (7.5s). Moreover, we observe that despite wrong initialization of $\frictionCoeffEst$ it rapidly converges into the region close to the ground truth value of tire/road friction coefficient $\frictionCoeff$.

\subsection{Impact of the dynamics model on the prediction accuracy}
    
\begin{figure}[t]
\begin{minipage}{0.48\linewidth}
    \centering
    \vspace{-0.2cm}
    \includegraphics[width=\linewidth]{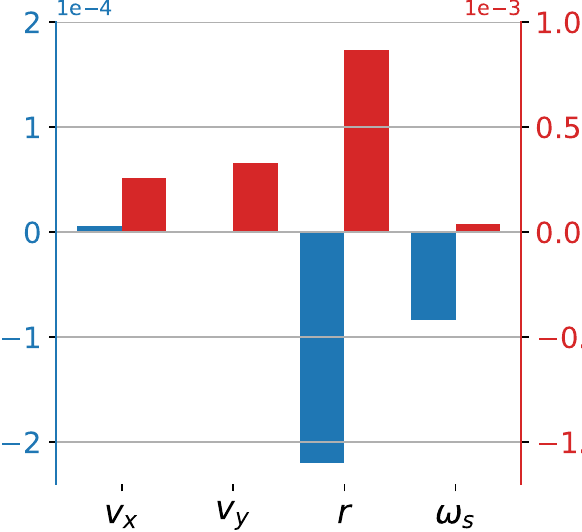}
    \vspace{-0.6cm}
    \caption{Difference of \gls{mse} between $\NNTireMu_{\UKF}$ and $\NNTireMu$ models in one-step prediction task (blue) and estimation task (red). Positive numbers indicates lower errors of $\NNTireMu_{\UKF}$ model.}
    \label{fig:pred_est_error_comp}
\end{minipage}%
\hspace{0.1cm}
\begin{minipage}{0.48\linewidth}
    \centering
    \includegraphics[width=\linewidth]{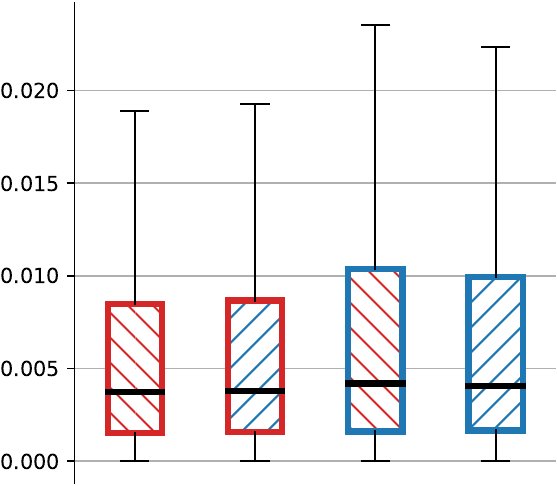}
    \vspace{-0.6cm}
    \caption{Impact of the dynamics model used in the prediction step (border) and observation model (hatching) on the squared estimation error. Red denotes the $\NNTireMu_{\UKF}$, while blue $\NNTireMu$.}
    \label{fig:pred_obs_model_test}
\end{minipage}
\vspace{-0.6cm}
\end{figure} 

To analyze the sources of the improved estimation performance while using dynamics models trained through \gls{ukf} we compared a $\NNTireMu_{\UKF}$ and $\NNTireMu$ on a one-step prediction and \gls{ukf} state estimation tasks. The results shown in Fig.~\ref{fig:pred_est_error_comp} indicate that the performance for the one-step prediction task achieved by $\NNTireMu_{\UKF}$ improved slightly for $\vx$, remained unchanged for $\vy$, and significantly dropped for $\yawRate$ and $\engineSpeed$.
This outcome may be attributed to the critical role of the $\vx$ in the vehicle dynamics and the fact that $\yawRate$ and $\engineSpeed$ are directly measurable. Interestingly, in the case of \gls{ukf} estimation task, using $\NNTireMu_{\UKF}$ instead of $\NNTireMu$ leads to significantly better estimation of all states. We attribute this result to the simultaneous optimization of dynamics and noise models while training through \gls{ukf}, which enables the dynamics model to focus on hard to measure states and allows for adequate noise model adaptation.

Moreover, due to the fact that the vehicle dynamic model is a part of both the prediction and update step, we tested which of them is responsible for the gain in estimation performance between $\NNTireMu$ and $\NNTireMu_{\UKF}$. To do so, we evaluated \glspl{ukf} for which the dynamics models used for prediction and update steps may be different and were either $\NNTireMu$ or $\NNTireMu_{\UKF}$ (see Fig.~\ref{fig:pred_obs_model_test}). One can see that from the estimation accuracy perspective, the use of the dynamics model trained through \gls{ukf} is particularly important in the prediction step and leads to notably smaller errors (18\% decrease in \gls{mse}).

\subsection{Impact of the training sequence length}
\label{sec:seq_len}

\begin{figure}[t]
\centering
\includegraphics[width=\linewidth]{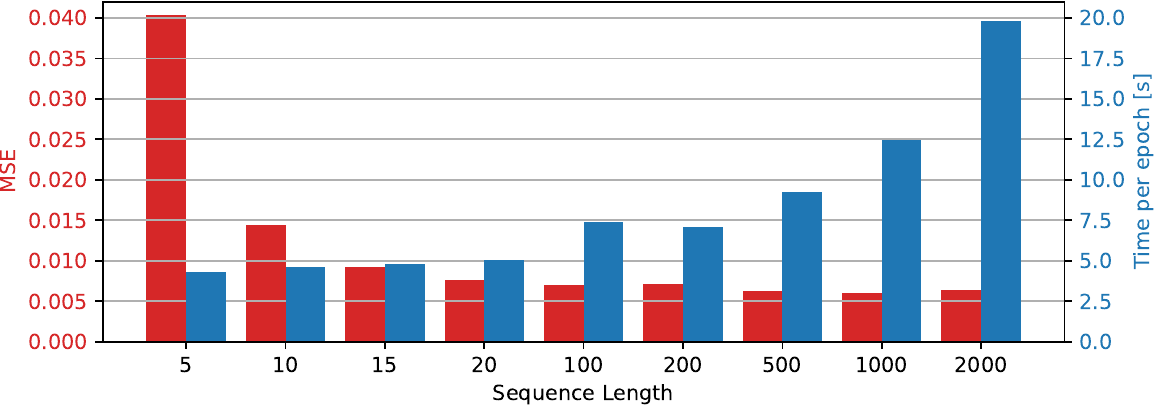}
\vspace{-0.6cm}
\caption{MSE and average training epoch duration (for batch size 256 using AMD Ryzen 9 5900X) as a function of training sequences length.}
\label{fig:ukf_sequence_len}
\vspace{-0.2cm}
\end{figure} 

One of the main parameters when training dynamics models through \gls{ukf} is the length of the training sequence. We analyze its impact on the estimation accuracy and training time in Fig.~\ref{fig:ukf_sequence_len}.
Similarly to \cite{kloss2021howtotrain} we observe that increasing training sequence length leads to decrease in estimation loss but at the same time lengthens the training.
In all previous experiments, we choose a sequence length of 500 samples as a compromise between training time and accuracy.

\section{CONCLUSIONS}

In this paper, we analyzed the impact of learning the dynamics model using \gls{ukf} on the velocity estimation performance in autonomous racing. We show, in contrast to previous studies~\cite{kloss2021howtotrain}
, that learning the dynamics model through a differentiable filter leads to significantly improved estimation performance. Moreover, we propose to extend the tire forces modeling approach presented in~\cite{xu2022tireforcesnn} with an online estimated tire-road friction coefficient. This approach not only outperforms the \gls{rnn}-based state estimators in the nominal conditions but also enables a zero-shot generalization to the significantly lower friction coefficient, which is not achieved by any considered state-of-the-art method.
We extended these results with an analysis of the sources of the improved prediction accuracy due to learning through \gls{ukf}, which we attribute to the possibility of simultaneous adaptation of the dynamics and noise models.
Finally, we introduced a publicly available dataset of aggressive maneuvers that covers a significantly larger range of conditions than existing datasets.

\bibliographystyle{IEEEtran}
\bibliography{ours}

\end{document}